\documentclass[10pt,journal,compsoc]{IEEEtran}

\usepackage{graphicx}%
\usepackage{multirow}%
\usepackage{amsmath,amssymb,amsfonts}%
\usepackage{amsthm}%
\usepackage{mathrsfs}%
\usepackage{xcolor}%
\usepackage{textcomp}%
\usepackage{manyfoot}%
\usepackage{booktabs}%
\usepackage{algorithm}%
\usepackage{algorithmicx}%
\usepackage{algpseudocode}%
\usepackage{array}
\usepackage{listings}%

\theoremstyle{thmstyleone}%
\theoremstyle{thmstyletwo}%
\newcommand{\Cov}{\mathrm{Cov}}
\theoremstyle{thmstylethree}%

\usepackage{algorithm,algpseudocode}
\newtheorem{prop}{Proposition}
\usepackage{xcolor}

\ifCLASSOPTIONcompsoc
  \usepackage[nocompress]{cite}
\else
  \usepackage{cite}
\fi
\ifCLASSINFOpdf
\else
\fi
\begin{document}
%
\title{Learning from Random Subspace Exploration: Generalized Test-Time Augmentation with Self-supervised Distillation}
%
%
%
%

\author{Andrei~Jelea,
        Ahmed~Nabil~Belbachir
        and~Marius~Leordeanu
\IEEEcompsocitemizethanks{\IEEEcompsocthanksitem All authors are with NORCE Research AS.
\IEEEcompsocthanksitem A. Jelea and M. Leordeanu are with "Simion Stoilow" Institute of Mathematics of the Romanian Academy.
\IEEEcompsocthanksitem M. Leordeanu is with "National University of Science and Technology "Politehnica" Bucharest.}
}

%
%

\markboth{
}%
{Jelea \MakeLowercase{\textit{et al.}}: Learning from Random Subspace Exploration: Generalized Test-Time Augmentation with
Self-supervised Distillation}
%



\IEEEtitleabstractindextext{%
\begin{abstract}
We introduce Generalized Test-Time Augmentation (GTTA), a highly effective method for improving the performance of a trained model, which unlike other existing Test-Time Augmentation approaches from the literature is general enough to be used off-the-shelf for many vision and non-vision tasks, such as classification, regression, image segmentation and object detection. By applying a new general data transformation, that randomly perturbs multiple times the PCA subspace projection of a test input, GTTA creates valid augmented samples from the data distribution with high diversity, properties we theoretically show that are essential for a Test-Time Augmentation method to be effective. Different from other existing methods, we also propose a final self-supervised learning stage in which the ensemble output, acting as an unsupervised teacher, is used to train the initial single student model, thus reducing significantly the test time computational cost. Our comparisons to strong TTA approaches and SoTA models on various vision and non-vision well-known datasets and tasks, such as image classification and segmentation, pneumonia detection, speech recognition and house price prediction, validate the generality of the proposed GTTA. Furthermore, we also prove its effectiveness on the more specific real-world task of salmon segmentation and detection in low-visibility underwater videos, for which we introduce DeepSalmon, the largest dataset of its kind in the literature.
\end{abstract}

}

\maketitle

\IEEEdisplaynontitleabstractindextext

%
\IEEEpeerreviewmaketitle

\section{Introduction}
\label{sec:intro}

Test-Time Augmentation (TTA) is a popular strategy used in a recent class of vision
methods which are effective for improving at test time the performance of a trained model. It works as follows: multiple versions of the input are passed through the same model and an ensemble is formed by the several output candidates produced, with an improved accuracy over the initial prediction. The main limitation of the existing techniques is that each one is particularly designed for a specific vision task, completely lacking generality. Given the effectiveness of these methods, there is real value in designing a general TTA method, which could be used off-the-shelf for various tasks in machine learning and that is the main point we propose in this paper. Our Generalised Test-Time Augmentation (GTTA) is general, has provable statistical guarantees and can be used as a single procedure for many vision and non-vision tasks, as we demonstrate in extensive experiments. 

In a nutshell, GTTA works as follows: the test input data is projected in the PCA subspace of the entire training data set, where it is randomly perturbed with Gaussian noise, multiple times, taking into account the variance explained by each subspace component. The resulting latent representative samples are reconstructed in the initial input space and passed through a previously trained model to form an ensemble of outputs, which are averaged in order to obtain the final prediction. By randomly exploring the natural subspace of the input data for the given task, we are guaranteed to produce automatically samples that are appropriate for that particular task, without the need to manually design different task-specific data augmentation and transformation procedures. Moreover, as we will show in the theoretical analysis section, the PCA projection along with the random exploration within the PCA subspace have additional statistical benefits which justify why our convenient, off-the-shelf approach is also highly effective in practice.

Traditional TTA methods are also slow at test time, as they require the passing of different versions of the input through the same model, multiple times. We address this limitation as well, by introducing a self-supervised learning strategy, in which the ensemble output is distilled into the initial base model on novel unlabeled data. Interestingly enough, the distilled single model ends up matching GTTA ensemble's performance, while retaining the lower single-forward pass cost at test time. The end result is that final self-distilled GTTA model is more general, better in terms of accuracy and significantly faster at test time than other TTA methods proposed for different tasks in the literature, which we compared against.

\subsection{Contributions} 

We make contributions along several directions: \\

\textbf{1) GTTA:} A general TTA approach that is highly effective for different vision and non-vision tasks, faster at test time (due to its final self-supervised distillation stage), more accurate than existing task-specific methods and with desirable theoretical guarantees. We experimentally demonstrate the superiority and generality of GTTA, compared to other TTA approaches and SoTA models, on different well-known tasks and datasets (CIFAR100, COCO, House Prices Prediction, X-Ray Pneumonia Detection, Age Group Prediction, Speech Recognition).\\

\textbf{2) GTTA has sound statistical properties:}
We proved the effectiveness of our proposed method with several theoretical contributions. Foremost, we showed, for the first time to our best knowledge, that under some mild assumptions, a TTA method is guaranteed to improve over the initial model. Then, we proved that a TTA method such as GTTA, which produces a large variance of candidates' outputs, is more efficient than others which do not, as a superior transformation diversity leads to a better improvement over the initial model. Finally, we study how we can choose GTTA hyperparameters in order to control the final estimator errors. We showed both theoretically and experimentally that GTTA has these desirable properties - which explains its advantage over existing methods in various real world experiments. We also introduce, based on our statistical analysis, an automatic procedure for selecting the optimal level of Gaussian noise to be applied in the PCA subspace for a particular test sample.\\

\textbf{3) Self-supervised distillation for reduced computational cost at test time:}
We distill GTTA ensemble output into the initial single model on new unlabeled data, in a self-supervised setting, with the single student model matching the performance of the ensemble teacher. Thus, the test cost remains that of a single model inference pass and the training cost requires only to retrain, a second time, the initial model. Existing TTA methods do not perform this final self-distillation stage, remaining costlier at test time. \\

\textbf{4) Effective uncertainty measure for improved self-supervised learning:} 
We introduce a novel uncertainty measure, based on that TTA ensemble variance, which is positively correlated with its actual errors - when many candidates in the ensemble say the same thing (low ensemble variance), the final ensemble output is more likely to be correct. This idea can be effectively used to downplay, in the self-supervised learning cost, the ensemble outputs (used as pseudolabels) with high variance (low consensus), leading to a better final performance.\\

\textbf{5) Novel dataset for fish segmentation in low-visibility underwater videos:} We noticed in various experiments the robustness of GTTA in low-quality images, a context in which the performance gap between GTTA and other methods is actually increasing. It seems that GTTA is comfortable in such difficult scenarios, a behavior that is statistically justified by the fact that Gaussian noise in the PCA subspace domain is actually the source of variation in the GTTA ensemble candidates, which
provides robustness to noise in the input. Given this observation, we tested GTTA on a specific real-world problem, relevant in aquaculture and marine industry, that of fish segmentation and counting, which is performed in difficult, poor visibility underwater videos. 

\textbf{DeepSalmon dataset:} Since methods that require a good image clarity, such those using optical flow~\cite{8,15}, do not work in underwater environments of limited visibility, the majority of existing approaches for fish segmentation rely on heavy supervised training~\cite{6,5,4,7}. However, there are only few available annotated fish datasets in the literature, such as DeepFish \cite{1}, Seagrass \cite{2}, for fish only, and YouTube VOS \cite{3}, with $94$ object categories, including fish. In order to address the limited labeled fish data in the literature, we introduce DeepSalmon, a relatively large video fish dataset of 30GB (see Fig.~\ref{fig:fig6} and~\ref{fig:fig7}), with $12$ difficult videos (at $25$-fps) of \emph{Salmon Salar} species in two built-in systems: a control tank and an ozone tank. Most of the fish in our dataset are hard to detect, even by human eye, due to the poor visibility, delusive appearance of the environment and large number of fish that appear and occlude each other. We provide annotations at both semantic and instance levels for $200$ video frames. Due to the difficulty of the task, it took about $40-60$ mins to fully annotate a single frame. Note that the limited underwater optical view makes it impossible to effectively use label propagation methods for automatic annotation. Compared to the other few existing datasets in the literature, DeepSalmon captures difficult and harder-to-solve underwater fish scenarios, due to significantly larger number of fish in the videos, which also have poorer visibility.

Furthermore, we also address the task of fish counting, for which we introduce a novel segmentation-based object counting approach (Sec. \ref{sec:fish_counting}) that significantly surpasses the SoTA YOLOv8 model on DeepSalmon dataset and it benefits as well from the proposed Generalized Test-time Augmentation (GTTA) method.

\subsection{Related work}

\indent \textbf{Related work on Test-Time Augmentation:}
Our approach belongs to a relatively recent class of ensemble methods, that of Test-Time Augmentation (TTA), which aggregates predictions across different augmented versions of a test input image, to form a final ensemble output, in order to improve models' performance at test time for tasks such as image      classification~\cite{38,39,40,son2023efficient,kandel2021improving}, object detection~\cite{42,43,44,gonzalo2021improving} and semantic segmentation~\cite{45,46,47,wang2018test,nalepa2019training}. However, all these methods are designed for particular vision tasks, using different vision-specific augmentation functions, such as color jittering~\cite{52,53}, cropping and scaling~\cite{49,50} or rotation and flipping~\cite{46, 56}. 

In contrast, GTTA applies a transformation that is truly general across tasks and domains, by perturbing with random Gaussian noise the PCA projection of a test input
on the natural subspace of the training data for that particular task.
Moreover, the existing methods are not robust to poor quality data and do not provide any statistical analysis and theoretical justification of their performance.
Only in~\cite{38} we find some insights about the effects that certain augmentations have on data when performing TTA, but the type of augmentations (image cropping and scaling) are again specific to vision. Also, different from previous TTA approaches, we provide an effective way to learn self-supervised, by distilling the output of the ensembles into the initial single model, while the training cost is weighted by a novel uncertainty measure 
computed based on ensemble candidates' output variance, a measure which again can be statistically justified.

For clarity, it is worth mentioning that Test-Time Augmentation strategy is different from Test-Time Adaptation~\cite{liang2025comprehensive, karmanov2024efficient, niu2024test, chen2024each} class of methods, also abbreviated as TTA, which refers to the process of adapting a pre-trained model's parameters to a new, unseen target domain during inference. In contrast to Test-Time Adaptation techniques, where the model should be adapted for every new inference data, GTTA creates a robust ensemble and distills its output on unlabeled data, different to the actual test data, improving in this way model performance and generalization, while keeping computational test time at a single model inference pass, with just one model re-training phase.  
However, it is also worth observing that the self-supervised distillation
phase of our GTTA could also become a test-time adaptation scheme if the distillation is performed on the ensemble output of test samples.
\\

\textbf{Related work on Autoencoding:} Learning efficient data representations has been approached in many works from the literature. Autoencoders are a particular class of these methods, where an encoder compresses data into a latent space representation and a decoder reconstructs the input. Examples are classical methods like K-means~\cite{66} and Denoising Autoencoders (DAE)~\cite{67} or more recent works such as Variational (VAE)~\cite{68,76,77,78} and Masked Autoencoders (MAE)~\cite{69, 79,80,81}. 

Different from other Test-Time Augmentation techniques, GTTA uses an Autoencoder to automatically project test input samples into the natural latent space of the given task-specific data and, after applying Gaussian noise for the compressed representation, to reconstruct it in the initial input space. As Autoencoder we use Principal Component Analysis (PCA)~\cite{66}, because is fast, general enough to be used for any task and domain. Moreover, the importance of each subspace component is directly obtained, a property that is effectively used by GTTA.\\

\textbf{Related work on Uncertainty Estimation:}
Understanding model predictions is crucial in many applications for improving aspects such as safety, trustworthiness and decision making process. Examples of recent works on uncertainty estimation in the literature are~\cite{74}, which introduces an uncertainty-guided mutual consistency learning framework for semi-supervised medical image segmentation and~\cite{75}, where the uncertainty, which is estimated as Kullback–Leibler divergence between student and teacher models’ predictions, is used to rectify the learning of noisy pseudo-labels.

Different from other existing methods, GTTA ensemble obtains the consensus across multiple augmented versions of a test input, produced by randomly perturbing its PCA subspace projection, and further distills the output into the initial model, using a novel self-supervised strategy, which estimates predictions' correctness based on within-ensemble output variance. Our observation, which is intuitive and also provable under mild statistical conditions, is that the higher the consensus among GTTA ensemble's candidate outputs, the more likely it is that ensemble output (as average over all candidates) will be correct. We integrate 
this intuition into the self-supervised learning cost, by weighing more the pseudo-lables (produced by the ensemble teacher) which have a higher level of ensemble consensus (lower uncertainty).\\

\textbf{Related work on Self-supervised Distillation:} 
By combining self-supervised learning~\cite{84,85,86} with a knowledge distillation strategy~\cite{87,88,89}, self-supervised distillation methods~\cite{90,91,92,93} aim to enhance the performance of a model, leveraging the rich representations learned by a teacher to guide a student model, facilitating in this way efficient learning. 

Different from existing Test-Time Augmentation methods in the literature, in order to reduce the testing cost and improve generalization as well, we take a self-supervised teacher-student learning approach by distilling the output of GTTA ensemble into the single initial model. As mentioned before, we weigh the self-supervised loss function with our novel proposed variance-based uncertainty measure.

Note that, while there are other approaches that train a single student model on pseudo-labels given by an ensemble teacher~\cite{croitoru2019unsupervised}, they use ensembles of 
$N$ \textit{distinct} models, with different set  of parameters, which is the traditional way followed by most ensemble methods~\cite{20, 21, 22, 25, 24,82,83}. In our case,
we train the student with an ensemble teacher using the exact same model with the same parameters for all candidates.

\section{Generalized Test-Time Augmentation}
\label{sec:approach}

Often the causes of estimation errors in machine learning are due to subtle
but systematic and structured noise present in the data. The real-world task of fish segmentation in low-quality, poorly illuminated underwater videos, which we also tackle in our experiments, is a good example of a task where positive signal (e.g. fish), could be easily confused with background clutter (plants, shadows and other structures that usually appear in underwater images). If we could find a way to wash out the distracting and structured clutter from the data, model prediction process will be simplified and improved. This specific real-world task, posing
several challenges brought by difficult filming underwater conditions is one source of inspiration for creating
Generalized Test-time Augmentation. The other motivation comes from the widespread need in artificial intelligence for a more general 
approach that is less sensitive to structured or non-structured noises, can learn with limited resources, such as: limited memory and 
training computational cost, and very scarce
labeled  training data. As we will show in more details in the following pages, all these limitations are addressed by GTTA: 
\textbf{1)} it is general and can be effectively applied to many tasks; \textbf{2)} it is robust to noise, as it uses an effective way to create ensembles, with provable statistical properties; \textbf{3)} it requires low memory and computational cost, as a single model is trained for creating ensembles 
of many candidates and \textbf{4)} it can improve the initial model to generalize from limited labeled training data, by self-supervised distillation of the ensemble output. 

As mentioned, GTTA works as follows: given a trained model, at test time, the input data is projected onto the PCA subspace of the entire training set, where we then apply random Gaussian noise for the latent representation and the resulting noisy sample is reconstructed in the initial input space. We repeat the procedure multiple times to obtain a pool of candidate outputs, for a single given test input, to form a test-time ensemble whose output (as average over all candidates) is generally superior to the initial single-model output. As a final stage, during a self-supervised learning procedure, the GTTA ensemble acts as a teacher, on novel unlabeled data, for the base pretrained single-model (initially trained in a supervised way). The random exploration in the PCA subspace is so general that GTTA can be used off-the-shelf in any learning task, from any domain with real-valued data. 
This a main advantage over all the other task-specific TTA methods in the literature. 
Additionally, the last self-learning phase offers a test-time speed advantage for GTTA over other TTA approaches, by distilling the ensemble power into the initial single model.

Now we present in detail every step of our proposed GTTA method. The initial test input $\mathbf{I}$ is projected onto the PCA subspace, which is computed for the entire training set: $p_i = (\mathbf{I} - \mathbf{I}_0)^\top \mathbf{u}_i$ where $\mathbf{u}_i$, $i \in [0, \ldots, n_u]$ are the principal components and $n_u$ depends on the task, as explained in the Section \ref{sec:exp}.
Then \textbf{noise} is sampled independently from a Gaussian distribution $\mathcal{N}(0,\,\sigma^{2}_i)$ for every component and added to $p_i$: $p_i' = p_i + \textbf{noise}$. \\

\noindent We consider \textbf{two strategies for choosing the noise level} in our approach.

\begin{enumerate}
\item Use a constant noise level (= standard deviation of Gaussian noise independently sampled along all dimensions in the PCA subspace), added multiple times to a given test sample, to form the ensemble of output candidates. We introduce noise for the component $i$ with $\sigma_i = \sqrt{\lambda_i} \cdot \sigma$, where $\lambda_i$ is the eigenvalue corresponding to component $i$, obtained by projecting a test sample in the PCA subspace of the training set and $\sigma$ is a hyperparameter that controls the level of noise.
\item Use different noise levels for every candidate in the ensemble. We apply an incremental \textit{std} policy where we add noise for the $j$-th candidate in the ensemble with $\sigma_{i} = \frac{\sqrt{\lambda_i} \cdot (j - 1)\cdot \sigma}{N}$, where $N$ is ensemble size and $\sigma$ controls again the level of noise.
\end{enumerate}

Finally, the noisy latent sample is reprojected in the initial space: $\mathbf{I}' =  \mathbf{I}_0 + \sum_{i=1}^{n_u} p_i' \mathbf{u}_i$ and the augmented input is passed through the pre-trained model.
In Algorithm~\ref{alg:algorithm1} we summarize the steps of our approach, when using an incremental \textit{std} strategy.

\begin{algorithm}[t]
  \caption{Generalized Test-Time Augmentation (GTTA)}\label{alg:algorithm1}
  \begin{algorithmic}[1]
    \renewcommand{\algorithmicrequire}{\textbf{Input:}}
	\renewcommand{\algorithmicensure}{\textbf{Output:}}
    \algnewcommand\INPUT{\item[\algorithmicinput]}
    \algnewcommand\algorithmicx{\textbf{b) Semi-Supervised Procedure}}
    \algnewcommand\INPUTT{\item[\algorithmicx]}
    \Require
      \Statex Previously trained \textbf{Model}
      \Statex Ensemble size \textbf{N} and noise level  \textbf{$\sigma$}
      \Statex Test input input data sample \textbf{I} and training set \textbf{T}
    \Ensure
      \Statex GTTA prediction for the sample \textbf{I}
    \State Apply PCA on \textbf{T} and get principal components: $\mathbf{u}_i$, $i \in [0, \ldots, n_u]$
    \State Project \textbf{I} on every component: $p_i = (\mathbf{I} - \mathbf{I}_0)^\top \mathbf{u}_i$
    \For{$j = 1$ to $\textbf{N}$}        
        \State $\sigma_{i} = \frac{\sqrt{\lambda_i} \cdot (j - 1)\cdot \sigma}{N}$
        \State Generate \textbf{noise} $ \sim \mathcal{N}
        (0,\,\sigma_i^{2})$
        \State $p_i' = p_i + \textbf{noise}$
        \State $\mathbf{I}' =  \mathbf{I}_0 + \sum_{i=1}^{n_u} p_i' \mathbf{u}_i$
        \State $\textbf{predictions}(i,:) = \textbf{Model}(\mathbf{I}')$
    \EndFor
    \State $\textbf{finalPred} = $ mean$(\textbf{predictions}$, axis = 0$)$
    
  \end{algorithmic}
    \end{algorithm}
    
\subsection{Self-supervised Distillation of the GTTA Ensemble}

\begin{figure}[]
\begin{center}
   \includegraphics[width=0.8\linewidth]{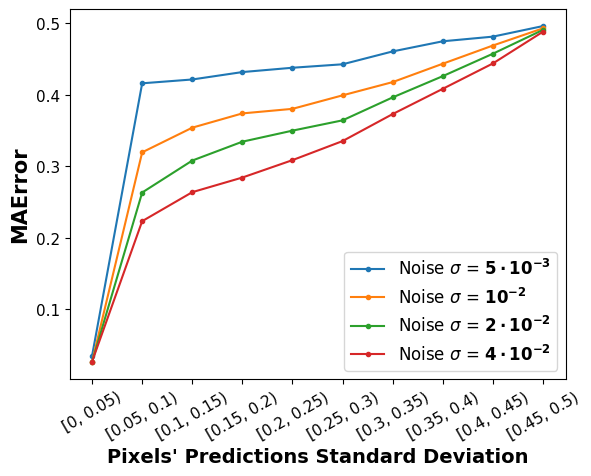}
\end{center}
   \caption{Relationship between the standard deviation (measure of variation, inconsistency) among the ensemble candidate outputs per pixel and their mean absolute error, with respect to ground truth, on DeepSalmon test set, for multiple noise levels added to the input sample (using the first noise adding strategy). The plot clearly shows, for all noise levels, that the higher standard deviation in the outputs (which can always be measured at test time), the higher true error (which is not known at test time) will be. Or, conversely, the stronger the consensus among candidates, the better the output. Based on this observation, we will use the standard deviation as a measure of certainty, that is of trust in the ensemble output - which is effective for self-supervised learning where the ensemble acts as a teacher for the initial single-model student.}
\label{fig:pix}
\end{figure}

In order to reduce testing cost and possibly improve generalization,
we use the output of GTTA ensemble as an unsupervised teacher, for new unlabeled data,
to retrain the base single model. As labels we use the initial ground truth for the supervised learning training data and the pseudo-labels produced by GTTA ensemble for an additional
unlabeled data set (which will not be used for final evaluation and testing). \\

\textbf{Novel measure of uncertainty for self-supervised ensemble distillation.} Through this self-distillation process, the initial model learns from the more powerful, robust ensemble. Moreover, the ensemble offers additional information regarding its own uncertainty, as explained next: during experiments, we observe a strong correlation between the standard deviation in ensemble output and the actual ensemble error. In Fig.~\ref{fig:pix} we show the relation between the standard deviation of the ensemble outputs, per pixel, and expected true error at that particular pixel from our semantic segmentation experiments on our DeepSalmon dataset (Sec. \ref{exp:DeepSalmon}). We tested for different levels of noise levels used to create the ensemble and observed the same strong correlation. The conclusion is clear: the smaller the standard deviation in GTTA ensemble output (that is, the stronger the consensus among ensemble candidates) the lower its true error. Therefore, the standard deviation of the ensemble outputs, which can be easily computed at test time, can act as a proxy for uncertainty, which is not known at test time. The higher the standard deviation (lack of consensus), the higher the expected error will be, that is, the higher the uncertainty of the GTTA ensemble output (which acts as the unsupervised teacher).  

Consequently, we use the standard deviation of the ensemble output as a \textbf{measure of uncertainty} and construct a weighted self-supervised learning cost, in which per-feature ensemble output samples with lower standard deviation (lower uncertainty), are more important as pseudo-labels than per-feature outputs with higher standard deviation:

\begin{equation}
    L(p, y) = - \frac{1}{\sum_{i} w_{i} } \sum_{\substack{i=1 \ldots M}} w_{i} y_{i} \log p_{i},
\end{equation}
where $w_{i} = 1 - s_{i}$ and $s_{i}$ is standard deviation value in the ensemble outputs for feature $i$, always smaller than $1$ and $M$ is the number of features.
This novel weighted semi-supervised cost proves highly effective in our experiments, leading to a better final performance (see, for example, Fig. \ref{fig:jj}).

\section{Intuition and Statistical Analysis}
\label{sec:analysis}

Despite being an effective method for improving the performance of a trained model, there is insufficient discussion in the literature regarding the theoretical aspects of Test-Time Augmentation. 


\subsection{GTTA reduces the initial model's error}

Below we present our main theoretical result for GTTA, which is that it is expected to lower the initial model's error. Before we present the actual proposition and its proof, we first motivate the main statistical assumption on which it is based: we assume that
the individual GTTA transformations (PCA projection + random perturbation)
do not change the expected error of the model's output. 
The GTTA transformations $\mathcal{T}_i$ represent a projection on the PCA subspace of the input data distribution $D$, followed by a random perturbation within that distribution (along the principal components and proportional to the standard deviation along each specific component) around the initial input sample $\mathbf{x} \sim D$. This means that the generated samples, as transformations of $\mathbf{x}$, are expected to be valid samples from the input data distribution and of similar difficulty with the original input $\mathbf{x}$. That is because they are randomly sampled around $\mathbf{x}$ from the same training distribution, which the model is familiar with. Therefore, it is reasonable to expect that the model will make a similar amount of error for the transformed samples $\mathcal{T}_i(\mathbf{x})$ as for the initial $\mathbf{x}$. 

We argue that, in fact, all TTA methods implicitly make this assumption, that the transformed samples are of
similar difficulty with the initial samples, when they apply various geometric transformations (e.g. scaling and rotation) and appearance changes (e.g. changing the illuminant or other color transformations). While none of the other works have provided a solid theoretical reason for why TTA improves performance, we show in this Section that our PCA-based augmentation approach is more general and explores more fully the space of transformations which do not change the error (the level of difficulty of the data samples). This could also open a door for even more powerful TTA methods, based on the observations made here. Next we provide our main theoretical result. \\

\noindent \textbf{General TTA's Error Correction Property:}

\begin{prop}
Let $\mathcal{T}$ be the set of GTTA transformations (augmentations) 
$\mathcal{T}_i \in \mathcal{T} \;  i\in \{1 \dots n\}$ applied to an input sample 
$\mathbf{x} \sim D$ at test time and $\epsilon(f(\mathbf{x}))$ be the error with respect to ground truth for any output function $f$, given the input $\mathbf{x}$. We make the assumption that, for a given model $m$, the GTTA transformations do not change the expected squared $L2$ model error: $\mathbb{E}_{\mathbf{x} \sim D}[\epsilon(m(\mathbf{x}))^2] = \mathbb{E}_{\mathbf{x} \sim D}
[\epsilon(m(\mathcal{T}_i(\mathbf{x})))^2]$. 
Then GTTA is expected to reduce the model's initial error: $\underset{x \sim D}{\mathbb{E}}[\epsilon(\frac{1}{n}\sum_{i=1}^{n}m(\mathcal{T}_i(\mathbf{x})))^2] \leq \underset{x \sim D}{\mathbb{E}}[\epsilon(m(\mathbf{x}))^2]$. 
\end{prop}

\textbf{Proof:}
For any input $\mathbf{x}$, the model's output can be written as $m(\mathbf{x}) = y^* + \epsilon(m(\mathbf{x}))$, where $y^*$ is the ground truth and the second term is the signed error.

Then  GTTA ensemble output, the average model's outputs over all applied transformations $\mathcal{T}_i$, is:
\begin{equation}
\frac{1}{n}\sum_{i=1}^{n}m(\mathcal{T}_i(\mathbf{x})) = \frac{1}{n}\sum_{i=1}^{n} (y^*  + \epsilon(m(\mathcal{T}_i(\mathbf{x}))) =  y^* + \frac{1}{n}\sum_{i=1}^{n}\epsilon_i,
\end{equation}

where $\epsilon_i$ is the signed error (continuous or discrete) corresponding to the model output for transformation $\mathcal{T}_i$. 
Note that, while we specifically work with a scalar output model, the theorem and its proof can be immediately extended to vector outputs.

From the equation above, it follows that the error of the GTTA ensemble output for a given $\mathbf{x}$ 
is:  $\epsilon_{GTTA}(\mathbf{x}) = \frac{1}{n}\sum_{i=1}^{n}\epsilon_i$.
Then, using the sum form of Jensen's inequality, given that the squared L2 error function is convex, we have: 

\begin{align}
\epsilon_{GTTA}(\mathbf{x}) = (\frac{1}{n}\sum_{i=1}^{n}\epsilon_i)^2 \leq \frac{1}{n}\sum_{i=1}^{n}\epsilon_i^2 
\end{align}

Thus, the squared error of the GTTA ensemble for a given input sample $\mathbf{x}$ is less than (or equal) to the average squared errors of the individual outputs for each input augmentation.

Since the above equality is true for any input $\mathbf{x}$, it is also true for the expected values of the left and right hand sides, which will conclude our proof:

\begin{align}
\epsilon_{GTTA}^2 =
\mathbb{E}_{\mathbf{x} \sim D} [(\frac{1}{n}\sum_{i=1}^{n}\epsilon_i)^2]
& \leq \mathbb{E}_{\mathbf{x} \sim D}[\frac{1}{n}\sum_{i=1}^{n}\epsilon_i^2] \\
& \leq \frac{1}{n}\sum_{i=1}^{n}  \mathbb{E}_{\mathbf{x} \sim D}[\epsilon_i^2] \\
& \leq \mathbb{E}_{\mathbf{x} \sim D}[\epsilon^2].
\end{align}

\qed \\

\textbf{Discussion:} We note that the above result becomes a strict inequality if the errors are not identical, which is the case almost always in practical experiments - the model's outputs are not identical due to the random independent Gaussian noise added, therefore the signed errors are different.

In conclusion, the theorem says that if the augmentations preserve the "structure" of the model's output, without increasing the expected error, while keeping the data in the same valid space as the initial model's output, then a TTA method reduces the model's error. 

GTTA  proposed here is expected to be highly effective in reducing the error, 
due to the following points:
\begin{enumerate}
\item Data augmentation based on the PCA projection onto the natural subspace of the training data does not introduce additional errors, as it keeps the augmented samples within the same data distribution on which the model was trained.
\item The Gaussian noise added to each PCA component independently increases the diversity of the augmentation and increases their total variance, which has the effect of making the inequality stronger (
increasing the Jensen Gap), thus making GTTA more effective - a property discussed in the next Section \ref{sec:Jensen_Gap}
\item The experiments validate the theoretical properties. For example, in image segmentation tests on the DeepSalmon dataset (see Figure \ref{fig:deacreasing_bias}), we observe that as we increase the noise level the GTTA error decreases significantly w.r.t to the initial model (when noise is zero), until the error reaches a minimum. After that point the error starts increasing as the stronger noise destroys the valuable "signal" in data and the augmented samples depart from the natural data distribution.
\end{enumerate}

To our best knowledge, in the literature there is only one theoretical result~\cite{kimura2021understanding},
which discusses statistical properties of TTA. The authors show that the error of TTA depends on the ambiguity of the output, without providing specific assumptions to guarantee TTA's error correction. Furthermore, no experimental results are presented in~\cite{kimura2021understanding} and no specific TTA method is proposed.

In the next Sections we provide and discuss additional statistical properties of GTTA, to better justify the points made above.

\subsection{GTTA is effective in error correction}
\label{sec:Jensen_Gap}

In this Section we show theoretically why, under some mild and inituitive assumptions, GTTA is expected to be highly effective in reducing the error over the initial model, by producing valid augmented samples which have a high diversity and total variance. The result below addresses this property, and it basically says that a TTA method producing samples of high total variance (for a given input) is expected to be better than one producing samples of lower total variance.
\\
\noindent \textbf{GTTA's Efficiency Property:}

\begin{prop}
\textbf{a)} GTTA Gaussian random noise de-correlates the data along the different PCA subspace dimensions.
\textbf{b)} If we assume that for a given prediction model $m$ when the total variance of the input data increases, the total variance of the output also increases, that is $\mathrm{Var}[\mathbf{x_1}] > \mathrm{Var}[\mathbf{x_2}] \Rightarrow \mathrm{Var}[m(\mathbf{x_1})] > \mathrm{Var}[m(\mathbf{x_2})]$, then: as the total variance of augmented samples produced by a TTA method for a given input increases, the final the TTA error decreases - as long as the TTA method produces samples
for which the model $m$ is expected to have the same error as for the initial sample (assumption of Proposition 1).
\end{prop}

\textbf{Proof of part a):} 
Since we add independent noise with $0$ mean and $\sigma_i$ standard deviation for the PCA subspace components, each feature $i$ (we sample values at dimension component $i$),
in the PCA subspace has the distribution $\mathcal{N}(p_i,\,\sigma^{2}_i)$, where $p_i$ is initial feature value.

If we consider 2 components $i, j$ with the distributions $X_i = \mathcal{N}(p_i,\,\sigma^{2}_i)$ and $X_j = \mathcal{N}(p_j,\,\sigma^{2}_j)$, then $\Cov[X_i,X_j] = \Cov[p_i + \mathcal{N}(0,\,\sigma^{2}_i), \ p_j + \mathcal{N}(0,\,\sigma^{2}_j)] = \Cov[\mathcal{N}(0,\,\sigma^{2}_i), $ $ \  \mathcal{N}(0,\,\sigma^{2}_j)] = 0$, as we sample noise independently for every component. Since $\mathrm{Var}[X_i] = \sigma^{2}_i \; \forall i$, therefore the covariance matrix for features' distributions in PCA subspace have a diagonal form,  with the eigenvalues equal with $\sigma^{2}_i$, for all principal components $i \in [0, \ldots, n_u]$. \\

\textbf{Proof of part b):} 
Let $T_1$ and $T_2$ be two Test-Time Augmentation (TTA) methods that satisfy the assumptions of Propositions 1 and 2, such that $\mathrm{Var}[\mathbf{x}_{T_1}] > \mathrm{Var}[\mathbf{x}_{T_2}]$. It follows that:
\begin{equation}
    \mathrm{Var}[m(\mathbf{x}_{T_1})] > \mathrm{Var}[m(\mathbf{x}_{T_2})].
\end{equation}

We have: $m(\mathbf{x}_{T_1}) = \mathbf{y^*} + \epsilon(m(\mathbf{x}_{T_1})$, where $\mathbf{y^*}$ is the ground truth and $\epsilon(m(\mathbf{x_{T_1}}))$ is the corresponding error. Then, the error variance of TTA method $T_1$ is:

\begin{equation}
    \mathrm{Var}[\epsilon(m(\mathbf{x_{T_1}}))] = \mathrm{Var}[m(\mathbf{x}_{T_1}) - \mathbf{y^*}] = \mathrm{Var}[m(\mathbf{x}_{T_1})]
\end{equation}
and, similarly:
\begin{equation}
    \mathrm{Var}[\epsilon(m(\mathbf{x_{T_2}}))] = \mathrm{Var}[m(\mathbf{x}_{T_2})]
\end{equation}

From Eq. 7,8 and 9 above, it follows that:
\begin{equation}
    \mathrm{Var}[\epsilon(m(\mathbf{x}_{T_1}))] > \mathrm{Var}[\epsilon(m(\mathbf{x}_{T_1}))], 
\end{equation}
which says that $T_1$ has a larger error variance, while (and this is important) 
having the same expected model error for its samples $\mathbf{x}_{T_1}$ as $\mathbf{x}_{T_2}$ (assumuption from Proposition 1).

In previous work~\cite{simic2008global}, authors show that a larger total sample variance leads to a large Jensen gap (the Jensen inquality becomes stronger). In our case the samples are the individual errors for individual candidate samples produced by a TTA method. While these errors have the same expected value (assumption of Proposition 1) for each TTA method, $T_1$ and $T_2$, the variance of model errors for $T_1$ is larger than the variance of model errors of $T_2$ (Eq. 10 above). Therefore, for the same convex error function (L2 norm), the same expected individual candidate sample error, but a larger Jensen gap~\cite{simic2008global}, the TTA method T1 is expected to have a lower overall error of the ensemble output: $\epsilon_{T_1}(\mathbf{x}) < \epsilon_{T_2}(\mathbf{x})$ \qed \\

\textbf{Discussion:} The first part of Proposition 2 shows that GTTA de-correlates the TTA transformed data samples along the fundamental PCA dimensions of the data, and by doing so it effectively increases diversity and
influences the total variance of the TTA samples. By considering the correct maximum number of representative PCA dimension, GTTA in fact maximizes the total variance, while still satisfying the first assumption (Proposition 1), which is to not increase the error of the output after a single transformation. Then, the second part of the result, shows that a TTA method with a higher total variance is expected to perform better, while the two assumptions are met (Propositions 1 and 2). Note that it is important to distribute the noise along the different data subspace dimensions, for a given total variance, in order to not fall outside the natural data distribution by putting too much variance (noise) along a single dimension. Thus, the two assumptions encourage GTTA to perform well against other TTA method that limit the transformations within a space with much fewer degrees of freedom, such as color jittering, cropping and others. 

On the other hand, by using the natural number of dimensions provided by PCA and adding noise along those dimensions, GTTA provides both high total variance and diversity (independence of the samples given the input). Thus GTTA can take full advantage of the conclusions of Propositions 1 and 2, which explains its better performance in practice.

In Table \ref{table:var} we compare the total variance at input (data samples produced by the TTA transformations) and model output (after applying the model to the transformed samples) for GTTA and two highly popular augmentation techniques: color jittering, which implies changing brightness, contrast, saturation and hue in the images and AugMix \cite{63}, a highly effective method which applies multiple augmentation functions to a image, including translation, solarization, rotation and also color jittering. We augment every image from DeepSalmon test set $N = 100$ times using the optimal hyperparameters values, according to ground truth, for each of the 3 methods, and we compute top $30$ eigenvalues of the sample covariance matrix. Next, we repeat the procedure for the outputs obtained after passing each transformed version of the input data through the model, for each TTA method. In case of GTTA we use the same standard deviation $\sigma_i$ for all components. Finally, we compute the normalized total variance at both input and output levels for all Test-Time Augmentation methods as the sum of covariance matrices' eigenvalues. Note that GTTA has a much better input data transformation diversity compared to the other 2 TTA methods, which leads to a significantly higher variance at the output level, as well.  In light of Proposition 2, This result could explain experimentally why GTTA outperforms in our experiments existing augmentation methods.

\subsection{GTTA removes structured and systematic noises}

Here we discuss another interesting property of GTTA.
By decorelating the data transformations along the sub-space data dimensions, GTTA manages to "destroy" structured, systematic noises present in the initial data input. In some sense, this behavior is expected when random independent noise is added along orthogonal data dimensions. Random independent Gaussian noise destroys all structures, but when the quantity of this noise is not two large and when noise is added within the natural data subspace, the variation induced does not destroy the useful "signal" but only the structure of "inconvenient" structured noises in the original data.

\begin{figure}[]
\begin{center}
   \includegraphics[width=0.8\linewidth]{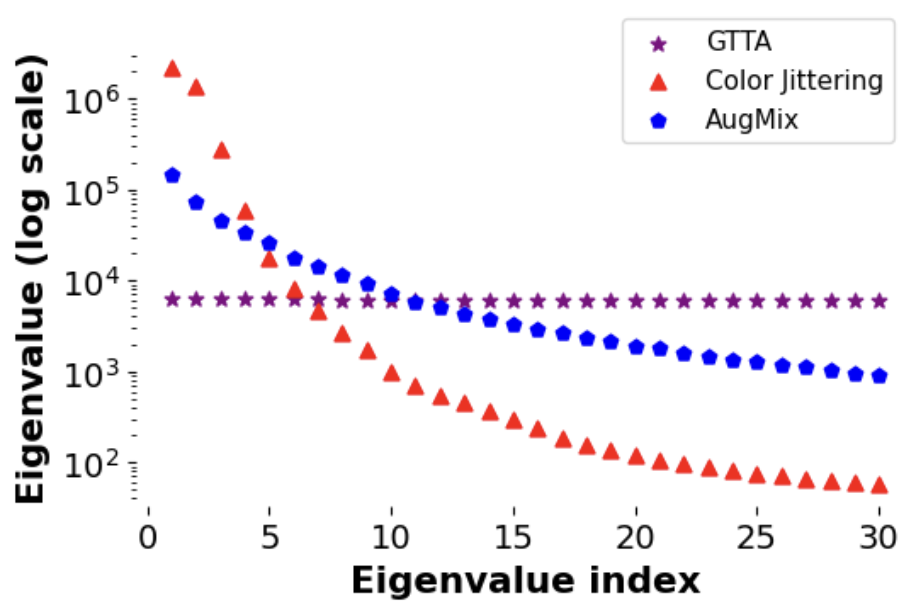}
\end{center}
   \caption{Top $30$ eigenvalues of the sample covariance matrix over DeepSalmon test set for GTTA, color jittering and AugMix methods. Number of samples is $N = 100$. Note how the candidates produced by GTTA are the most uncorrelated and thus, diverse. The inter-dependence of the other TTA methods is due to the fewer degrees of freedom of those respective transformations, that automatically results in a less diverse population of candidates, in which the structure noises have better chances to survive. For example, color jittering, which is defined by a few global parameters for the entire image cannot destroy a specific shape in the background clutter, while GTTA, with its purely random noise in the class subspace can. We apply noise equally to each component in our approach.}
\label{fig:eig}
\end{figure}

In Fig. \ref{fig:eig} we experimentally validate this property of GTTA, comparing it to color jittering and AugMix. Once again we compute the eigenvalues of the sample covariante matrix at input level over DeepSalmon test set, but in order to demonstrate the "power" of random noise we construct this time the covariance matrix for our method in the PCA subspace, using the noisy latent samples when equal noise levels were added to each component.
Figure \ref{fig:eig} illustrates the average of the eigenvalues over the entire DeepSalmon test set. Note how in case of our method the eigenvalues are equal, which highlights the diagonal form of the covariance matrix and shows why GTTA produces a strong augmentation diversity.

\begin{figure}[]
\begin{center}
   \includegraphics[width=\linewidth ]{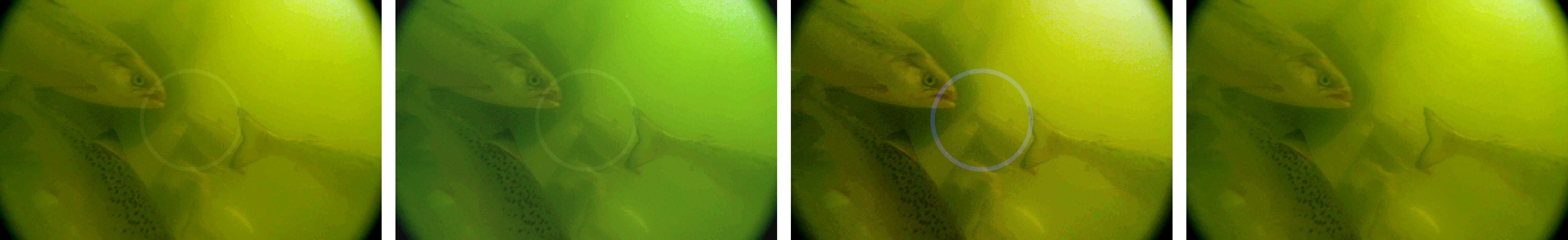}
   \hspace{5.2em} \textbf{a)} \hspace{5.2em} \textbf{b)}  \hspace{5.2em}   \textbf{c)} \hspace{5.2em}   \textbf{d)}
\end{center}
   \caption{Examples of augmented versions of a test image from DeepSalmon dataset (shown in \textbf{a}), with a manually inserted structural distractor in the form of a circle, as produced by three different TTA methods: 
    \textbf{(b)} color jittering, \textbf{(c)} AugMix, \textbf{(d)} GTTA. Note that only GTTA removes the added structured noise.}
\label{fig:circle_noise}
\end{figure}

In order to better illustrate experimentally the property of GTTA to remove structured and systematic noises from the initial data input, we performed the following controlled experiment. We manually introduced noise in the form of a circle in an image from the DeepSalmon dataset, then applied the previous Test-Time Augmentation techniques: color jittering, AugMix and GTTA. To better emphasise the robustness of GTTA, we introduced the same structural noise in half of the images from the training set, which are used for creating the PCA subspace. Fig.~\ref{fig:circle_noise} shows how GTTA washes out completely the structure of the distracting circle, while the circle can be clearly seen in the augmented versions produced by the other two TTA methods.

\subsection{Concluding remarks on why perturbing data in PCA subspace is effective}

We generally identify two main causes of classifier or regression error:
\begin{enumerate}
\item Changes in input structure that are representative for the semantic class of interest, but cannot be correctly recognized by the model, due to insufficient training (e.g. a car seen from a novel point of view or a slightly different type of car not seen during training).
\item Changes in input that are due to structured, systematic noises, such as clutter, occlusions, shadows and other distractors. Such noises could be subtle, but their structured appearance can lead to estimation errors.
\end{enumerate}

\noindent GTTA addresses both of these challenges:
\begin{enumerate}
\item To tackle the first cause of error, we create multiple input versions in the natural subspace of the data, learned unsupervised using Principal Component Analysis (PCA) on the training set: thus, we expect the random samples to be valid and representative of the data. For example, if we learn the PCA subspace for human faces, we expect that by randomly producing samples in the subspace around a particular input face, we will produce proper variations of valid human faces, probably similar to the initial input. These samples act as an effective data augmentation scheme, which is easy to produce at test time and is also general and applicable to virtually any type of data.
\item To address the second type of error, both the PCA projection and the addition of right amounts of independent Gaussian noise, for each component, after the PCA projection onto the subspace, have the combined desired effect of de-correlating the data along these components and increasing augmented data diversity. This effectively removes subtle but structured noises, which are unrelated to the given category/task of interest. 
\item Moreover, by increasing data diversity and total variance, while staying inside the original distribution, GTTA is also shown to enlarge the Jensen error gap to the original model, thus obtaining a significant correction of the final model error.
\end{enumerate}

In the next Section we will aim to study what is the optimal amount of noise to be added in order for GTTA to attain maximum efficiency.

\begin{table}[]
\caption{Input and output total variance produced by GTTA and other TTA methods, computed over DeepSalmon test set. Note how our method obtains a better diversity at both levels.}
\centering
\begin{tabular}{|c|c|c|}
\hline
Method & Input variance & Output variance \\
\hline\hline
Color Jittering \ & \ 0.3243 & \ 0.0072 \ \\
AugMix \ & \ 0.7543 & \ 0.0313 \ \\
GTTA \ & \ \textbf {3.6543} & \ \textbf {0.0678} \ \\
\hline
\end{tabular}
\label{table:var}
\end{table}

\subsection{Finding the optimal GTTA hyperparameters}

Next we look at how the level of noise ($\sigma$ value), ensemble size ($N$) and \textit{std} strategy can affect the final GTTA ensemble expected error.
We consider, for a particular test input and a $\sigma$ value, the statistical population of predictions generated by applying random noise with the fixed magnitude $\sigma$ to that input in the PCA subspace, then passing the reconstruction through the same pre-trained model. The population of outputs can be represented as a random variable, $Y$ and the target is to estimate the ground truth for that input, that is $\mathbf{y}^*$

If we repeat de process $N$ times, independently, the predictions represent $N$ observations, corresponding to i.i.d. random samples $\mathbf{y}_i$.
We then estimate the target output by taking the ensemble mean, $\bar{\mathbf{y}} = \frac{\mathbf{y}_1+ \mathbf{y}_2+ \ldots+ \mathbf{y}_N}{N}$. For the case of an incremental \textit{std} strategy, the difference is that observations are no longer identically distributed, there will be one population of predictions for every level of noise.

Our goal is to study which are the hyper-parameters $N$ and $\sigma$ that minimize the ensemble error,
defined as $\mathrm{Error}[\bar{\mathbf{y}}] = \mathrm{Bias}^2[\bar{\mathbf{y}}] + \mathrm{Var}[\bar{\mathbf{y}}]$, where $\mathrm{Bias}^2[\bar{\mathbf{y}}] = (\mathbb{E}[\bar{\mathbf{y}}] - \mathbf{y}^*)^2$ and $\mathrm{Var}[\bar{\mathbf{y}}] = \mathbb{E}[(\bar{\mathbf{y}} - \mathbb{E}[\bar{\mathbf{y}}])^2]$. 

The following result says that if the number of transformed GTTA candidate samples becomes sufficiently large, then GTTA error can be approximated by the bias component. Thus, the larger the number $N$ of samples we produce, the smaller the error, for a given level of noise $\sigma$. It is intuitive that the larger $N$, the better accuracy we expect, but it also comes at the expense of inference cost. However, the increase in inference time cost can be canceled by the self-supervised learning step, when a single student model learns from the GTTA ensemble teacher. As our experiments on different vision and non-vision tasks show, the self-supervised student has test-time similar accuracy as GTTA.

\begin{prop}
If we make the assumption that the model's output variance is bounded over the test set, $|\mathrm{Var}[m(\mathbf{x})]| \le c, \forall \mathbf{x} \in D$, GTTA estimator errors can be approximated by the bias component for a large enough $N$ value, as the estimator variance goes towards zero. 
\end{prop}

\textbf{Proof.}  For simplicity of notation we use $\mathbf{y} = m(\mathbf{x})$. We prove the proposition for each sampling noise level strategy. \\
\textbf{a) Constant noise levels:} Let be $\bar{\mathbf{y}} = \frac{\sum_{i=1}^{N} \mathbf{y}_i}{N}$ ensemble estimator for a particular noise level $\sigma$, with $\mathbf{y}$ the corresponding random variable. Then, $\mathbb{E}[\bar{\mathbf{y}}] = \frac{\sum_{i=1}^{N} \mathbb{E}[\mathbf{y}_i]}{N} = \mathbb{E}[\mathbf{y}]$ and $\mathrm{Var}[\bar{\mathbf{y}}] = \frac{\sum_{i=1}^{N} \mathrm{Var}[\mathbf{y}_i]}{N^2} = \frac{\mathrm{Var}[\mathbf{y}]}{N} \le \frac{c}{N}$, as $\mathbf{y}_i$ are independent. Since $\mathrm{Var}[\bar{\mathbf{y}}] \to 0$. Then, for a given noise level $\sigma$ the GTTA ensemble error can be approximated with the bias, as $N$ increases. \\

 \textbf{b) Incremental noise levels:} Once again $\mathrm{Var}[\bar{\mathbf{y}}] = \frac{\sum_{i=1}^{N} \mathrm{Var}[\mathbf{y}_i]}{N^2} \le \frac{c \cdot  N}{N^2} = \frac{c}{N} \to 0 $
 when $N \to \infty$. Therefore, in this case we also have $\mathrm{Error}[\bar{\mathbf{y}}] \approx \mathrm{Bias}^2[\bar{\mathbf{y}}]$ for a large enough value $N$.
 
 \qed \\
 
\begin{figure}[]
\begin{center}
   \includegraphics[width=\linewidth]{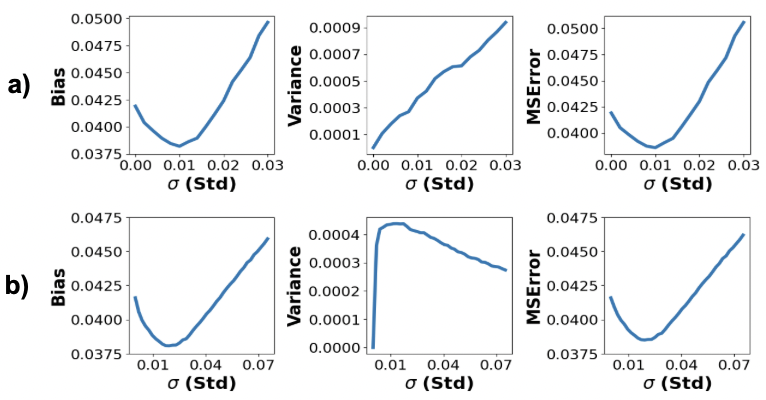}
\end{center}
   \caption{Estimator bias, variance and error evolution over DeepSalmon validation set for different noise level $\sigma$ values when a constant \textbf{(a)} or incremental \textbf{(b)} \textit{std} strategy is used for our GTTA method. Note how bias first decreases with larger \textit{std}. This indicates that a small amount of added noise is beneficial, as it has the ability to remove the potentially harmful structured noise in the data. As the amount of added noise increases over a threshold, it becomes too large and it starts destroying the good signal and structures in the data as well - those that are relevant for the given task and classes of interest. Also note that the variance is much smaller than the bias, and it can always be reduced towards zero by increasing the number of generated GTTA input samples.
}
\label{fig:deacreasing_bias}
\end{figure}

This result shows how, even though our method has a great input data augmentation diversity, the variance in the overall, final GTTA ensemble output (after averaging the candidates) can be easily reduced towards zero by increasing the number of candidates $N$. Then, the overal GTTA output error will be almost equal to the GTTA bias. Next we will see how the bias itself (thus, the total error for large $N$), depends on the noise level $\sigma$. We will see that as the sigma initially increases, the bias decreases up to an optimal point, from which the error starts increasing with the noise level. That observation makes perfect sense: initially, the noise introduced along the PCA components is beneficial, as we explore more and more inside the subspace. However, as the noise crosses a certain level, it becomes damaging, as it produces candidates which are no longer representative of the original data distribution in the PCA subsapce. More numerical details and discussions are presented next.

\subsection{Automatic estimation of optimal GTTA noise level}

In Figure \ref{fig:deacreasing_bias} we present the evolution of GTTA estimator bias, variance and total error over DeepSalmon validation set from our segmentation experiments in Sec.~\ref{exp:DeepSalmon}, when using constant or incremental levels of noise in our approach. As ensemble size we used $N = 15$ for the first strategy \textbf{(a)} and $N = \sigma$ for the second one \textbf{(b)}, in order to better cover the range of possible noise level values, $[0, \sigma]$. Using these ensemble sizes, the variance becomes insignificant compared to bias, which will now control the final errors in case of both strategies, as can be observed in Fig. \ref{fig:deacreasing_bias}. Interestingly enough, bias indeed is reduced as the \textit{std}
increases, until a minimum error is reached, which statistically justifies our GTTA approach. The optimal $\sigma$ values found for DeepSalmon validation set are $\sigma = 0.1$ and $\sigma = 0.2$, respectively, for the 2 strategies of our method. \\
 
We would like to be able to automatically determine the best $\sigma$ value for a particular test sample,
in the absence of ground truth at test time.

The core idea is based on the statistical properties we have discovered so far, from which we can draw two relevant insights:
\begin{enumerate}
\item On one hand, a larger noise level $\sigma$ is desired as it produces a more diverse pool of candidates, which offers more options for the final ensemble output, while also increasing the Jensen's Gap between the GTTA final error and the errors of the individual candidates.
\item On the other hand, we also want a strong agreement among candidates, which would indicate a higher accuracy of the final ensemble GTTA output (see Fig. 1). However, the two points seem to go against each other as a higher agreement is immediately obtained for free for a very small level of noise: if the added noise is zero, all candidates are identical so they are in perfect agreement. The question is how to put these two opposing forces together - which we explain next.
\end{enumerate}

\textbf{Automatic noise level estimation solution:} While being in apparent contradiction, the two points above can be naturally linked: the level of noise, as it tends to produce larger variance in agreement levels, makes it harder for candidates to agree, therefore a large agreement level is more significant and much harder to attain under a large noise level than under a smaller one. It seems that we need to rescale a given agreement level $a(\sigma)$ by a certain unit of measurement which is induced by the noise level $\sigma$. Using the training set distribution we can immediately evaluate that unit of measurement by estimating the empirical standard deviation of maximum agreement levels per sample $\sigma_a$ for a given task - note that this can be done in the same way regardless of the specific task, which could be either a classification or a regression problem. Regardless of which one it is, the agreement's peak for a given sample is computed simply as the peak of the normalized histogram of candidate's outputs in the pool. Then all such agreement peak levels are collected over the training set and their standard deviation $\sigma_a$ is estimated. That standard deviation can now be uitilized as a unit for measuring the significance of a given agreement on a test case. This is boils down to simply rescale that agreement peak $a$ by multiplying with the factor $\sigma_a$. Note that for a given training data distribution, these values depend on the noise level $\sigma$. Then, the algorithm for selecting the optimal noise level is simply finding the $\sigma^*$ for which the average of the rescaled maximum peak levels are maximized. In other words:

\begin{equation}
\sigma^* = \arg \max_{\sigma} a(\sigma)\sigma_a(\sigma)
\end{equation}


\section{Experimental Analysis}
\label{sec:exp}

We test GTTA on different vision and non-vision tasks and compare it, when available, with current TTA methods in the literature. Since there is no TTA published, to our best knowledge, for the non-vision tasks we tested on, we focused on showing how GTTA can boost the performance of the initial model. In the case of visual recognition and segmentation tasks, the experiments show that the effectiveness of GTTA as compared to other TTA methods is more pronounced when the input data is of low-quality, which is a desirable property especially in real-world cases, such as underwater imaging, where high quality images are not available.

\subsection{Classifying Low-resolution Images}
\label{exp:CIFAR}

We first compared our approach with other popular Test-Time Augmentation methods on image classification task, for which we used the well-known CIFAR100 dataset, with images of very low resolution (32x32). As base prediction model we used a ViT-Base Transformer~\cite{60}, pre-trained on ImageNet-21k~\cite{62} and fine-tuned on CIFAR100, with a final accuracy score of $89.94$ on the test set. For adding the noise, we employed a constant \textit{std} strategy. In all approaches compared we used $N = 15$ as ensemble size and we apply our proposed uncertainty-based procedure of selecting the optimal hyperparameters per test image for each method. The number of principal components for GTTA PCA subspace, $n_u$, is chosen so that they explain $k = 99 \%$ (value validated on a small set) from the total variance.
In Tab.~\ref{table:exp1} we show the relative percentage accuracy error change, compared to the base model, obtained by all tested TTA methods.
Note that our simple GTTA is the only one able to improve the accuracy of the base model, having a better final performance than other TTA approaches, including AugMix and Intelligent Multi-View TTA \cite{70}, which are complex data transformations that combine multiple strong augmentation techniques. Note also how our final semi-supervised student surpasses the performance of the pre-trained model as well, obtaining a final score close to GTTA ensemble teacher.

\begin{table}[]
\caption{Experiments on Image Classification for 3 datasets: CIFAR100, FairFace and Chest-Xray: Relative percentage accuracy error change, compared to ViT-Base model, produced by GTTA and other TTA methods. Best results are shown in bold, second best are underlined. Lower is better.
}
\centering
\begin{tabular}{|l|w{c}{1.5cm}|w{c}{1.5cm}|w{c}{1.5cm}|}
\hline
\multirow{2}{*}{TTA Method} & \multicolumn{3}{|c|}{Relative Percentage Error Change (\%) $\downarrow$} \\

 & CIFAR100 & FairFace & Chest-Xray  \\
\hline\hline
Cropping & + 33.67 & + 6.22 & - 3.25\\
Perspective & + 28.72 & + 2.47 & - 2.44\\
Elastic & + 26.74 & + 7.11 & + 4.12\\
Rotation & + 12.78 & - 0.12 & - 8.23\\
AugMix & + 3.96 & - 3.04 & - 7.43\\
Color Jittering & + 2.97 & - 0.27 & - 2.15\\
Multi-View TTA  & + 1.23 & - 1.45 & - 5.45\\
\hline
Self-Sup Student & \underline{- 1.98} & \underline{- 3.81} & \underline{- 8.33}\\
GTTA Ensemble & \textbf{- 2.97} & \textbf{- 4.93} & \textbf{- 10.78}\\

\hline
\end{tabular}

\label{table:exp1}
\end{table}

\subsection{Classifying Medium-quality Images}
\label{exp:Faces}

Next, we tested our method on the task of image classification, when the input has a medium quality, using 2 datasets. The first one introduces the task of classifying people in groups of ages, based on their faces, using FairFace dataset \cite{karkkainen2019fairface}, with images of 224x224 resolution. The data is race balanced, the images were collected from YFCC-100M Flickr dataset \cite{thomee2016yfcc100m} and the individuals are classified in 8 groups of ages: $0-2$, $3-9$, $10-19$, $20-29$, $30-39$, $40-49$, $50-59$, $60-69$, and $\geq 70$. The second data set consists of a collection of chest X-ray images \cite{wang2017chestx} and the task is to predict if pneumonia is present in a specific patient. Since the input are radiographic images, the data has a limited quality, which is a perfect setup for testing our method. In case of both datasets we compared our approach with the same TTA techniques as in our last experiment, we use the same strategy for selecting the hyperparameters and as model we choose a ViT-Base Transformer as well. Table~\ref{table:exp1} shows the results. Note how again, for this new datasets and tasks, GTTA and also our final distilled student model surpass other TTA methods, producing a significantly reduction of relative accuracy error change, compared to the base pre-trained model.

\subsection{Segmenting Images from High to Low Quality}
\label{exp:COCO}

\begin{figure}[]  
\begin{center}
   \includegraphics[width=0.8\linewidth]{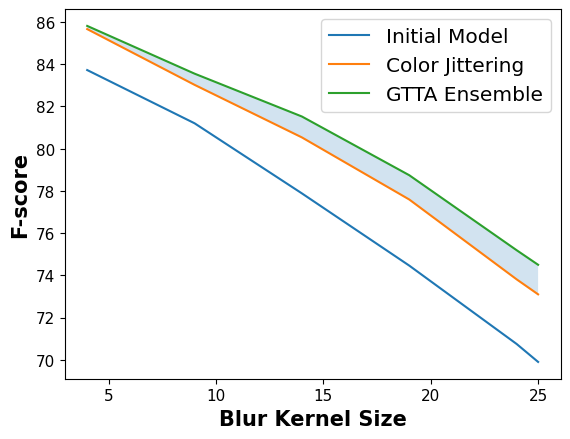}
\end{center}
   \caption{F-scores over blurred versions of test images from COCO dataset for initial Mask2Former model, GTTA and Color jittering TTA, using different levels of blur. Note how the GTTA advantage over color jittering increases as the image quality degrades.}
\label{fig:long}
\label{fig:coco}
\end{figure}

While we previously validated the effectiveness of GTTA on very low and medium resolution images, we did not yet study how effective GTTA is depending on specific levels of image quality. For this we selected the well-known COCO semantic segmentation dataset, with images of high-resolution and for which we can control the level of image quality (and implicit resolution) by varying the amount of image blur before applying TTA methods. Also, in the context of lower image quality, the task of image segmentation is perhaps more interesting, as it naturally requires a high level of attention to detail. For these tests we used as base model the State of The Art Mask2Former~\cite{61}, fine-tuned on COCO. We evaluated the performance with multiple levels of blur (box filter of different kernel sizes) and we choose for each method augmentations’ hyperparameters that obtain the best score for each test image. For this experiment, since the number of samples is smaller than the number of pixels in the image, we keep all the PCA components when applying GTTA. The results (Fig.~\ref{fig:coco})
show that GTTA initially outperforms color jittering, when the images were clear, and moreover, it is increasingly better as input degrades in quality, showing a much stronger robustness to low quality data.

\subsection{Non-Vision Tasks: Predicting House Prices} 
\label{exp:HousePrices}

In contrast to other augmentations, GTTA is general enough to be used for any task and domain. Thus, we also tested it on a completely different type of problem, that of predicting house prices, using the House Prices dataset from Kaggle competition. As base model we used a Multi-Layer Perceptron (MLP) with 3-hidden layers (of 256 neurons each), in order to predict house prices as real numbers. There are 36 numeric-valued inputs (e.g. LotArea, PoolArea) 
and 43 categorical ones (e.g. SaleCondition, LotShape), and we only constructed PCA subspace for the numerical variables. In order to ensure a small enough GTTA variance we chose as ensemble size $N = 100$ and we use $\sigma$ value and the total explained variance $k$ which obtains the best performance on the validation set. The results (Tab.~\ref{table:exp2}) show a relative error reduction of $3.15\%$ between the logarithms of true prices and GTTA predictions.

\begin{table}[]
\caption{House Price results: RMSError between the logarithms of true prices and predictions for base MLP model and GTTA. Note the benefit of using the off-the-shelf GTTA on a non-vision regression task and how the single-model self-supervised GTTA student performs similarly to the GTTA Ensemble, while having the much lower test-time inference cost (that of a single-model) as the Base MLP model. Lower is better.
}
\centering
\begin{tabular}{|c|c|}
\hline
Method & RMSError for log values of prices  $\downarrow$ \\
\hline\hline
MLP (Base model) \ & \ 0.1428 \ \\

GTTA Self-Sup Student \ & \ \underline{0.1378} \ \\
GTTA Ensemble \ & \ \textbf {0.1372} \ \\
\hline
\end{tabular}
\label{table:exp2}
\end{table}

\subsection{Non-Vision Tasks: Speech Recognition}
\label{exp:SpeechRec}

After testing our method on vision and tabular data, now it is time to show the capabilities of GTTA when used for other 2 popular types of input: language and audio data. For this, we select a task that bring together these 2 datatypes: automatic speech recognition (ASR), where the input spoken words and identified and converted into readable text. As base model we used the State-of-The-Art Whisper \cite{71}, a Transformer based encoder-decoder model trained on 680k hours of labeled speech data annotated using large-scale weak supervision. We evaluated GTTA on LibriSpeech dataset \cite{72}, a corpus of approximately 1000 hours of 16kHz English speech, derived from read audiobooks extracted out of LibriVox project \cite{73}. Whisper preprocesses the speech input by converting the audio frequencies to log-Mel spectrograms, which are then passed to the text transcription model. We apply noise in our method for the log-Mel representation of the audio input using a constant \textit{std} strategy and we use the hyperparameters (the number of principal components of the PCA subspace, $n_u$, and level of noise, $\sigma$) which obtain the best score on a validation set. Since GTTA candidates can have different lengths for this task, we keep only the generated text outputs in the ensemble with the same length (we choose the most frequent one) in order to be able the aggregate them based on predicted tokens' probabilities. The results (Table \ref{table:speech}) shows how GTTA improves the performance of the State-of-The-Art Whisper model for the task of speech recognition, reducing the relative error by $1.71 \%$.

\begin{table}[]
\caption{Speech Recognition results: Word Error Rate for initial Whisper model and GTTA. Again, GTTA is effective on improving over the initial base model. Note that the single-model self-supervised GTTA student performs similarly to the GTTA Ensemble, while having the much lower test-time inference cost (that of a single-model) as the Base Whisper model. Lower is better.
}
\centering
\begin{tabular}{|c|c|c|}
\hline
Method & Word Error Rate (WER)  $\downarrow$ \\
\hline\hline
\ Whisper (Base model) \ & \ 0.08056 \ \\
\ GTTA Self-Sup Student \ & \ \underline{0.07933} \ \\
\ GTTA Ensemble \ & \ \textbf {0.07901} \ \\
\hline
\end{tabular}
\label{table:speech}
\end{table}

\subsection{Difficult Real-World Environments: Fish Segmentation in Underwater Videos}
\label{exp:DeepSalmon}

\begin{figure*}[!ht]
\begin{center}
\includegraphics[width=0.85 \linewidth, , height=240pt]{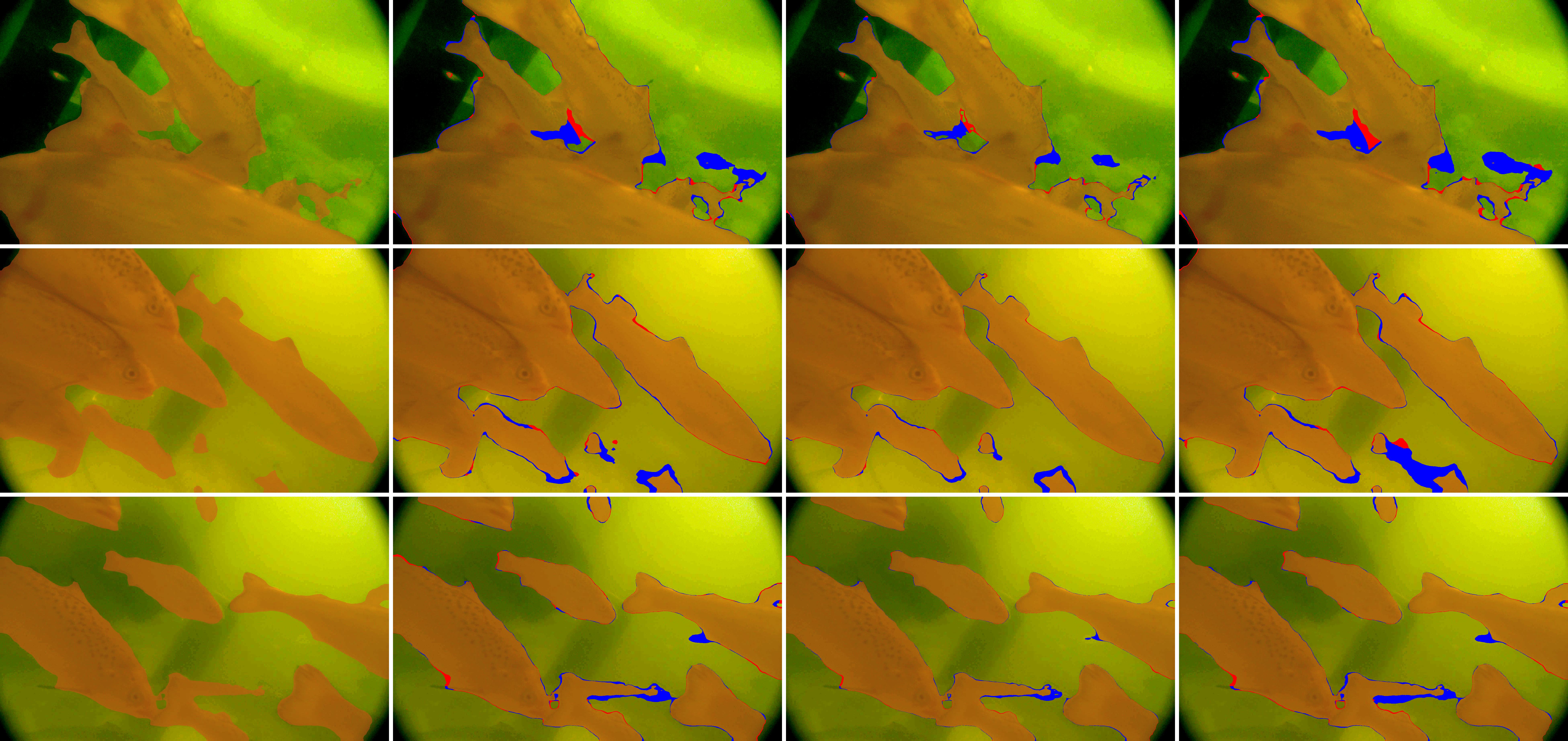}
\caption{Qualitative segmentation results on DeepSalmon. On each row, images shows in order results for:
\textbf{1st column:} base SegFormer model, \textbf{2nd column:} GTTA ensemble, \textbf{3rd column:} semi-supervised model
and \textbf{4th column:} weighted semi-supervised model. Pixel predictions where we differ from initial SegFormer model are shown in \textcolor{blue}{blue} (where we are correct) or \textcolor{red}{red} (where we are wrong). Over the test set, when we differ, we are correct (\textcolor{blue}{blue}) in more than 71\% of cases over the base model. Note that our methods, the GTTA and its self-supervised student, are consistently improving over the base model especially in difficult regions that lie at the object boundaries and capture fine shape details.
}
\label{fig:fig6}
\end{center}
\end{figure*}

Next, we performed experiments on our newly introduced underwater fish dataset. For the $200$ annotated images from DeepSalmon we use a 140-30-30 split for the train, validation and test sets, respectively. Each set comes from a different group of videos, 8 videos for training and 2 videos for validation and testing each. 

As base image segmentation model we used the State of The Art SegFormer, ver. b4 \cite{11}. For training SegFormer we used two approaches: \textbf{1)} fine-tuning the pre-trained model only using DeepSalmon dataset, and \textbf{2)} fine-tuning the model first on DeepFish and then on DeepSalmon. The results (Tab. \ref{table:t1}) show that the second procedure is better. Next we tested both strategies for injecting noise into the GTTA approach: $\textbf{1)}$ using a constant level of noise and $\textbf{2)}$ using an incremental noise magnitude for every new candidate. In both cases we apply the automatic procedure presented before for selecting the optimal $\sigma$ value that minimizes the uncertainty in the ensemble outputs and we keep all the PCA principal components.

\begin{table}[]
\caption{Maximum scores obtained on DeepSalmon by base SegFormer, with $(^*)$ and without an extra fine-tuning step on DeepFish dataset, baseline SegFormer ensemble (BEns.) and our GTTA ensemble, with a
constant (\textbf{ct}) or an incremental (\textbf{inc}) \textit{std} strategy.
Best results are shown in bold, second best are underlined. Lower is better.}
\centering
\begin{tabular}{|l|c|c|c|c|}
\hline
Method & F-measure & IoU & Precision & Recall \\
\hline\hline
SegFormer & 0.954 & 0.912 & 0.979 & 0.974 \\
SegFormer$^*$ & 0.958 & 0.920 & 0.983 & 0.978 \\
BEns. & 0.960 & 0.923 & 0.984 & 0.979 \\
GTTA Ens. (\textbf{ct}) & \textbf{0.964} & \textbf{0.930} & \underline{0.989} & \underline{0.982} \\
GTTA Ens. (\textbf{inc})& \underline{0.963} & \underline{0.928} & \textbf{0.990} & \textbf{0.984} \\
\hline
\end{tabular}

\label{table:t1}
\end{table}

We compared GTTA with the single SegFormer model and a standard ensemble formed by training (in the same way as the base model) 15 different SegFormer models and averaging the output maps (Tab. \ref{table:t1}). The results show the effectiveness of
GTTA for every metric considered (we report the maximum achieved by each method): Precision, Recall, F-measure, IoU. We observe that maximum precision and recall scores of GTTA are higher for an incremental standard deviation strategy, while IoU and F-scores are more similar, with a small plus for the constant \textit{std} strategy. Maximum metrics scores are computed over the whole Precision-Recall curve, by tuning the threshold applied on the final soft segmentation map.

Figure~\ref{fig:fig6} shows qualitative results. The majority of GTTA different pixel-level decisions from the base SegFormer model are accurate, GTTA helping at discovering previously unseen fish parts, especially in difficult regions of high uncertainty, near fish edges. This is a significant improvement in segmentation quality,
which is not correctly reflected by average performance values over the whole image, since such difficult regions, while very important, are relatively small in size. 

\subsection{Self-distillation for Fish Segmentation}
\label{sec:self-distillation}

\begin{figure}[]
\begin{center}
   \includegraphics[width=0.8\linewidth]{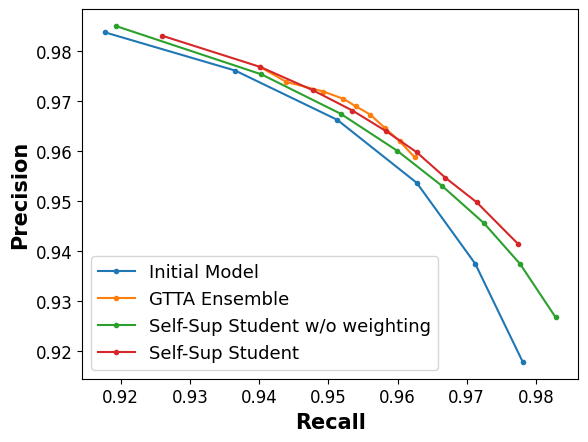}
\end{center}
   \caption{DeepSalmon experiments on semi-supervised learning: Precision-recall curves for the initial SegFormer model, GTTA ensemble and 
   the two variants of the GTTA self-supervised models, with and without our uncertainty-based cost weighting. Note how the PR curve of the weighted Self-Sup model matches the  GTTA teacher ensemble performance.}
\label{fig:jj}
\end{figure}

\begin{figure*}[!ht]
\begin{center}
\includegraphics[width=0.8\linewidth, height=250pt]{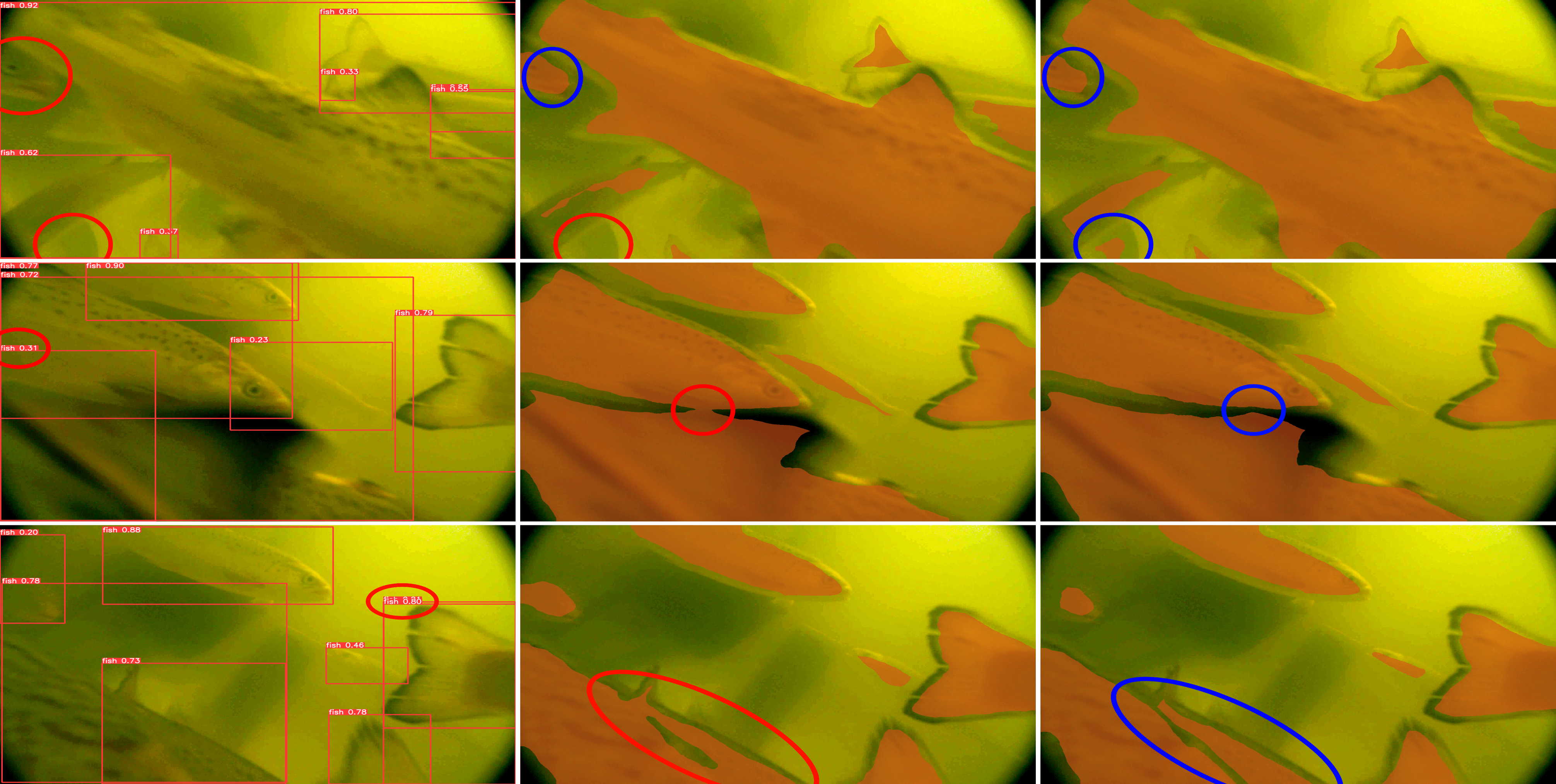}
\caption{Qualitative fish counting results on DeepSalmon. Each row illustrates predictions for \textbf{Left)} YOLOv8 model, \textbf{Middle)} our fish counting approach combined with our GTTA ensemble and \textbf{Right)} our weighted semi-supervised model. The errors made by these models are circled in \textcolor{red}{red} and their corrections are represented with \textcolor{blue}{blue}. Note how our methods detect barely seen fish that YOLOv8 misses.}
\label{fig:fig7}
\end{center}
\end{figure*}

Now we test the capability of GTTA ensembles to become unsupervised teachers, over unlabeled data, for the initial base model. For this task we extract from our DeepSalmon dataset $30$ frames per video, 50 frames apart. We use our GTTA method, with constant \textit{std} strategy (as in previous tests) to produce pseudo-labels for the newly extracted, unlabeled frames and then we distilled the base SegFormer model, pre-trained on the initial supervised set, on these pseudo-labels. Overall, this is a particular case of semi-supervised learning, in which the teacher-student system self-supervises itself, using the output of GTTA ensemble teacher to retrain the single, new generation student model.

\begin{algorithm}[t]
  \caption{SegCount Object Counting Method}\label{alg:algorithm2}
  \begin{algorithmic}[1]
    \renewcommand{\algorithmicrequire}{\textbf{Input:}}
	\renewcommand{\algorithmicensure}{\textbf{Output:}}
    \algnewcommand\algorithmicinput{\textbf{a) Training}}
    \algnewcommand\INPUT{\item[\algorithmicinput]}
    \algnewcommand\algorithmicx{\textbf{b) Testing}}
    \algnewcommand\INPUTT{\item[\algorithmicx]}
    
    \INPUT
    \Require

      \Statex Input frames $\mathbf{F_i}$
      \Statex Instance segmentation labels $\mathbf{L_i}$
    \Ensure
      \Statex Trained segmentation model for counting, \textbf{SegModel}
    \For{every training image $\mathbf{F_i}$}                    
    \For{every obj in instance segmentation map $\mathbf{L_i}$}       \State Extract segmentation mask $\textbf{S}_j$ for current obj
      \State $\textbf{erodedObjMask}_j = \textbf{S}_j \otimes \textbf{E}_1$
      \State $\textbf{T}_i = \textbf{T}_i + \textbf{erodedObjMask}_j$
    
    \EndFor
    \State Store training example $(\textbf{F}_i, \textbf{T}_i)$
    \EndFor
    \State Train segmentation model using training set $( \textbf{F}, \textbf{T})$

    \vspace{3pt}
        \hrule
    \vspace{3pt}
    
    \INPUTT
    \Require
      \Statex Test input frame \textbf{F}
      \Statex Trained segmentation model for counting, \textbf{SegModel}
    \Ensure
      \Statex Number of objects in frame \textbf{F}
    \State $\textbf{erodedSeg} = \textbf{SegModel}(\textbf{F}) \otimes \textbf{E}_2$
    \State Count the remaining connected components in $\textbf{erodedSeg}$
  \end{algorithmic}
\end{algorithm}

The plot in Figure \ref{fig:jj} shows Precision-Recall curves for the base SegFormer model, GTTA ensemble and our semi-supervised models.
Note how metrics scores are significantly increased for our
semi-supervised approaches, with the Weighted Self-Sup model (with variance-based pixel weighting) matching the GTTA ensemble teacher performance.
In Figure~\ref{fig:fig6} we present qualitative results which also show that our weighted semi-supervised model can match and often outperforms GTTA ensemble. The gain is enormous from a practical point of view, since the semi-supervised model has no additional test cost (compared to the initial base model), with only a small additional training one from fine-tuning the initial model on completely unlabeled data. 

\subsection{Fish Counting in Underwater Videos}
\label{sec:fish_counting}

\begin{table}[]

\caption{Fish counting results on DeepSalmon. Our final distilled student model surpasses GTTA teacher and reduces error by 33\%, compared to YOLOv8.}
\centering
\begin{tabular}{|l|c|}
\hline
Method & MAE score \\
\hline\hline
YOLOv8 & 1.8  \\
SegCount & 1.5  \\
SegCount + GTTA & \underline{1.3}  \\
SegCount + GTTA Self-Sup Student Model & \textbf{1.2}  \\

\hline
\end{tabular}
\label{table:t2}
\end{table}
We propose as our last contribution SegCount, a segmentation-based approach for object counting that can benefits as well from our GTTA method for performance improvement.
The idea of our approach is to predict smaller segmentation maps, corresponding to the interior of the objects, such that individuals will be well separated. 
In order to do this, we eroded each object independently in the instance level annotations, and trained a semantic segmentation model on these new eroded maps. At test time, after
a post-processing step, in which the output maps are also
eroded (in order to separate barely connected objects and
to remove small blobs), the number of remaining components represents the predicted number of objects. SegCount can be applied to counting any class of objects, depending on the available annotations. In algorithm \ref{alg:algorithm2} we summarized the steps of our segmentation-based object counting method.
We evaluate SegCount for fish counting task on DeepSalmon dataset using again a SegFormer as base segmentation model and we compare our approach with YOLOv8 (trained on the exact same images) by mean absolute error score (Table~\ref{table:t2}). Even without GTTA, our segmentation-based counting method outperforms YOLOv8 by a good margin. When SegCount is combined with GTTA, with or without semi-supervised distillation, the results are further significantly improved. Interestingly, SegCount with the weighted distilled single model outperforms SegCount with the GTTA ensemble.
Fig.~\ref{fig:fig7} shows interesting visual results, where YOLOv8 and SegCount+GTTA make some errors
(difficult cases of undetected, wrong-identified or overlapping fish), but SegCount+SemiSup model corrects them.

\section{Final Discussion and Conclusions}
We introduced GTTA, a highly effective and general Test-Time Augmentation method, which randomly explores
the natural subspace of the task-specific data to produce ensembles of output candidates, from input variations that are both representative for the given task and have less potentially harmful structured noise than those generated by other existing TTA approaches. These properties are justified by an in-depth statistical analysis and demonstrated on various vision and non-vision tasks and datasets. Different from other TTA methods, which are designed for specific vision tasks, another property of 
GTTA is its generality - the ability to be applied, off-the-shelf, with essentially no modification, to any given learning task.

GTTA is also versatile in a semi-supervised setting, through self-distillation, as demonstrated experimentally on several vision tasks, such as image segmentation and classification (in difficult scenarios, poor visibility and various image domains) and several non-vision tasks, such as predicting house prices and speech recognition. For the tasks of tasks of fish segmentation and counting in difficult, low-quality underwater vision, we also introduce the DeepSalmon dataset - the largest dataset for Salmon segmentation and counting. 

By distilling the GTTA ensemble into a single student model, our approach becomes fast at test-time
without a loss of performance. The effectiveness of the self-supervised learning procedure is further improved by a loss function in which the pseudo-labels are weighted according to an uncertainty measure, which is based on the standard deviation of the GTTA ensemble outputs. This novel measure of uncertainty is based on the intuitive insight, confirmed by empirical observation, that the higher the disagreement between the ensemble candidates (that is, the higher their standard deviation), the larger the ensemble's true error.

From a theoretical perspective, we showed, for the first time to our best knowledge, that under certain assumptions, a TTA method is guaranteed to improve over the initial model. We also proved that a TTA method such as GTTA, which produces a large diversity of candidates' outputs, is more efficient than others which do not. We showed both theoretically and experimentally that GTTA has these desirable properties - which explains its advantage over existing methods in various realworld experiments.

Moreover, GTTA presents a relatively high degree of robustness to low quality data, on different datasets and tasks (image classification on CIFAR, segmentation in COCO with various degrees of input quality), including the specific case of underwater fish segmentation and counting on the introduced DeepSalmon dataset. After testing GTTA along several dimensions, while changing the domain, tasks, the quality of the data and the amount of supervision, we can reliably validate the generality and reliability of our method. 

We believe that GTTA could open doors towards future TTA methods, in which the two main ideas proposed here could be pushed further: \textbf{1)}
data augmentation can automatically adapt to different types of domains and tasks, a strategy that can be very effective for TTA if it is also fast during test-time. This is the case with our PCA projection followed by random exploration. Future strategies could design other fast generative AI models, unlike the slow generative models of today, and \textbf{2)} unsupervised self-distillation of ensembles can be very effective if used in combination with powerful uncertainty-based measures for automatically weighing enerated pseudo-labels.

\ifCLASSOPTIONcompsoc
  \section*{Acknowledgments}
\else
  \section*{Acknowledgment}
\fi

This work is supported in part by projects “Romanian Hub for Artificial Intelligence - HRIA”, Smart Growth, Digitization and Financial Instruments Program, 2021-2027, MySMIS no. 334906; EU Horizon project ELIAS (Grant No. 101120237); “Unleashing the Sustainable Value Creation Potential of Offshore Ocean Aquaculture” (Grant No. 328724) and “The balancing act: Biologically driven rapid-response automation of production conditions in recirculating aquaculture systems (RAS)” (Grant No. 320717), co-funded by the Research Council of Norway.

\ifCLASSOPTIONcaptionsoff
  \newpage
\fi



%
\bibliographystyle{IEEEtran}
\bibliography{bibliography}

%

\begin{IEEEbiography}[{\includegraphics[width=1in,height=1.35in,clip,keepaspectratio]{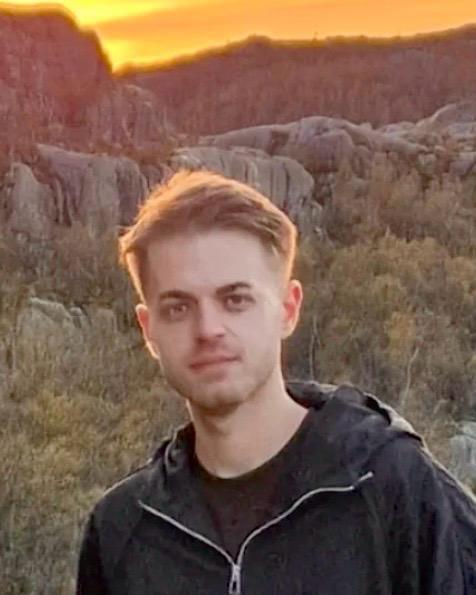}}]{Andrei Jelea}
is a doctoral student in computer science at the Institute of Mathematics of the Romanian Academy. His work is focused on unsupervised learning in images and videos, including self-supervised learning for image segmentation and object category discovery. Andrei obtained a Masters degree in Artificial Intelligence from University Politehnica of Bucharest (graduated in 2024) and two undergraduate degrees obtained with highest marks from two different universities and studied at the same time: Engineering Degree in Computer Science from University Politehnica of Bucharest (2022) and Bachelor Degree in Mathematics, from University of Bucharest (2021). During college and high-school Andrei was highly passionate about mathematics and computer science, obtaining several prizes at International and National Oympiads and Contests.
\end{IEEEbiography}
\vfill

\begin{IEEEbiography}[{\includegraphics[width=1in,height=1.35in,clip,keepaspectratio]{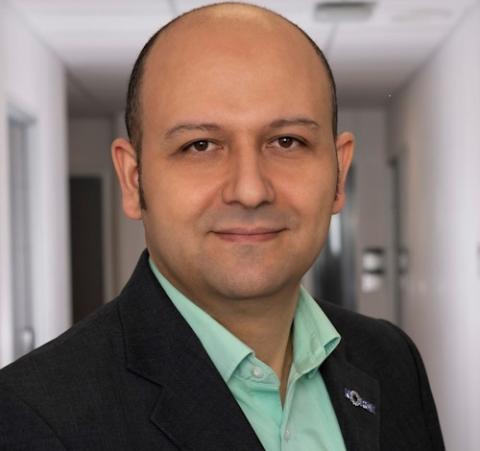}}]{Ahmed Nabil Belbachir} has almost three decades of experience with computer vision, AI and robotics, as engineer, scientist, visionary leader and strategist, contributing with sustainable innovative concepts, products and solutions. He is a research director of DARWIN group at NORCE, director at eu-robotics Aisbl and strategy advisor for ADRA Aisbl and the European Commission, for coordinating European Strategic Research, Innovation and Deployment Agenda and contributing to HEU work programme 2025-2027 in AI, Data and Robotics. He has a PhD (2005) in computer science from TU Vienna (Austria). He edited the single-source Springer book “Smart Cameras (2009)”, translated into Chinese by China Machine Press (2014) and has over 140 scientific publications, 3 patents. Actually, he is coordinating COGNIMAN, and iBot4CRMs, two Horizon Europe Innovation projects on AI-powered robotics, respectively for Flexible Manufacturing and for recovery of Critical Raw Materials  from electronic waste. He is also leading the Agder CARM centre developing novel robotics solutions on urban mining.
\end{IEEEbiography}
\vfill

\begin{IEEEbiography}[{\includegraphics[width=1in,height=1.35in,clip,keepaspectratio]{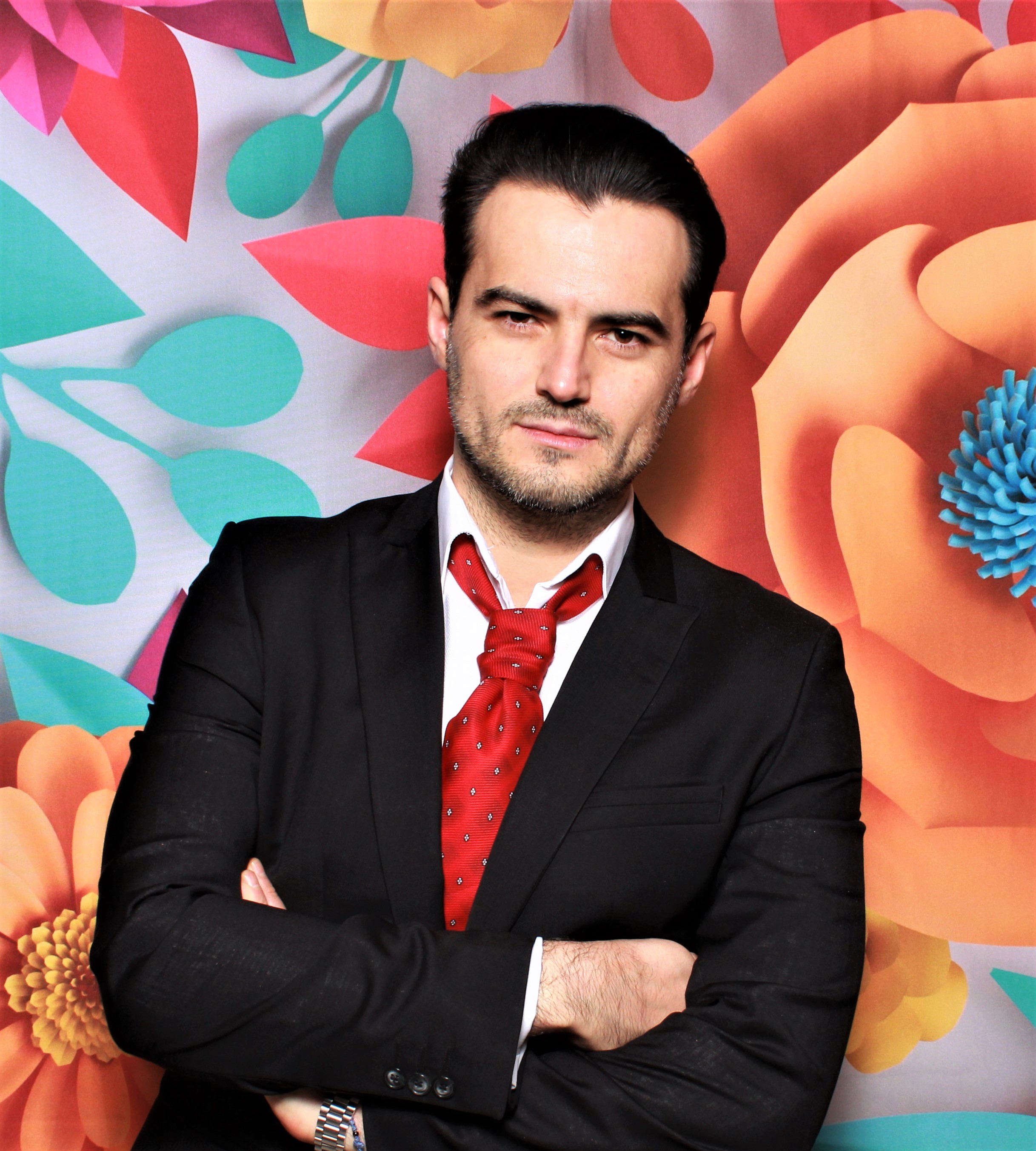}}]{Marius Leordeanu}
obtained a PhD degree in Robotics/Computer Vision from The Robotics Institute at Carnegie Mellon University (USA, 2009) and Bachelor's in Computer Science and Mathematics from Hunter College - CUNY (USA, 2003). Currently he is Full Professor at Polytechnic University of Bucharest and Senior Scientist at Norway Research Center (NORCE). His research focuses on different theoretical and practical aspects of unsupervised learning in space and time, with applications to several domains, including multi-modal and multi-task learning, video understanding, self-flying drones,  vision-language and human-AI interaction. Dr. Leordeanu won several prizes including Joseph A. Gillet Memorial Prize in Mathematics (USA, 2003), Computing Research Association (CRA) Outstanding Undergraduate Award (USA, 2003) as well as the (absolute) first prize at the National Physics Olympiad in Romania (1994). He also won as Principal Investigator eleven international and national research grants and in 2025 became one of the few Team Leaders of HRIA - Romanian Hub for Artificial Intelligence, the only one of this kind in Romania, gathering the top seven universities in the country. HRIA is the result of a five year effort, for which Marius Leordeanu was the initial founder and writer of the initial proposal draft. In 2024 Dr. Leordeanu was awarded by the Romanian Government the top prize in Computer Science and Mathematics at the Romanian Research Gala, while in 2014 he was awarded by the Romanian Academy the top prize in Mathematics, the “Grigore Moisil Award”, for his work on unsupervised learning and matching for graphs. He regularly serves as Area Chair for top conferences (CVPR, ICCV, ECCV, IJCAI, AAAI, WACV) and is currently Associate Editor for TPAMI. His book “Unsupervised Learning in Space and Time” (Springer, 2020) quickly became a best-seller on the topic of unsupervised learning.
\end{IEEEbiography}
\vfill



\end{document}